# Leveraging Knowledge Distillation for Efficient Deep Reinforcement Learning in Resource-Constrained Environments

Guanlin Meng
Department of Computing
The Hong Kong Polytechnic University
Hongkong, 999077, China
20099185d@connect.polyu.hk

*Abstract*—This paper aims to explore the potential of combining Deep Reinforcement Learning (DRL) with Knowledge Distillation (KD) by distilling various DRL algorithms and studying their distillation effects. By doing so, the computational burden of deep models could be reduced while maintaining the performance. The primary objective is to provide a benchmark for evaluating the performance of different DRL algorithms that have been refined using KD techniques. By distilling these algorithms, the goal is to develop efficient and fast DRL models. This research is expected to provide valuable insights that can facilitate further advancements in this promising direction. By exploring the combination of DRL and KD, this work aims to promote the development of models that require fewer GPU resources, learn more quickly, and make faster decisions in complex environments. The results of this research have the capacity to significantly advance the field of DRL and pave the way for the future deployment of resource-efficient, decision-making intelligent systems.

## I.  Introduction

Deep Reinforcement Learning (DRL) has emerged as a powerful technique for solving complex decision-making problems, with impressive results achieved in domains ranging from gaming to autonomous driving. DRL algorithms iteratively learn a strategy that maximizes cumulative rewards by dynamically interacting with the environment. Although they have achieved remarkable success in many applications, DRL algorithms face high computational requirements and slow learning rates, which pose significant challenges for their deployment in resource-constrained environments and real-time applications [1].

On the other hand, Knowledge Distillation (KD) is a popular machine learning technique that is used to transfer knowledge from larger, more complex models (known as teachers) to smaller, simpler models (called students) [2]. It is a highly valuable technique as it enables students to learn from the experiences of their more advanced teachers, thus transferring information that can be used to develop more accurate predictions. KD has been effectively used for a variety of purposes, particularly in reducing the computational and memory requirements of deep learning models without significantly affecting their performance [3].

DRL and KD are both popular techniques in machine learning today. However, the intersection of DRL and KD remains relatively unexplored. There is relatively little research on the effects of distillation on different DRL algorithms and the impact of varying distillation methods on DRL algorithms' performance. Despite their respective successes, it is still not clear how to best combine DRL and KD to achieve even better results in complex decision-making problems. A key question in this area is how to effectively distill the knowledge from a teacher DRL model to a student DRL model, taking into account the specificities of each algorithm. Answering this question could lead to more efficient and effective DRL algorithms that are better able to handle resource-constrained environments and real-time applications.

This paper aims to explore the applicability of KD techniques in the realm of DRL. By combining the power of DRL with the knowledge transfer capabilities of KD, researchers can unlock the potential for faster training, reduced computational resources, and effective decision-making in complex environments.

The integration of DRL with KD allows for the distillation of complex, high-performance reinforcement learning strategies into more straightforward and more efficient models. This synergy offers several benefits. Firstly, it enables the development of more efficient DRL models that are better suited for deployment on resource-constrained devices. By distilling various DRL algorithms and analyzing their distillation effects, this research provides a benchmark for evaluating the performance of different DRL algorithms refined using KD techniques.

Secondly, the incorporation of KD accelerates the learning process of DRL models, resulting in reduced training time and cost. This acceleration is crucial for achieving faster decision-making in complex environments, which is particularly relevant in real-time applications such as robotics, online recommender systems, and self-driving cars. By distilling DRL algorithms and exploring the combination of DRL and KD, this study endeavors to develop models that require fewer GPU resources, exhibit faster learning rates, and make quicker decisions in complex environments. Through this study, it is expected to shed light on the unexplored synergies between DRL and KD, paving the way for more efficient and resource-friendly AI algorithms.

## II. METHOD

### A. Environment

The environment used in this study is CartPole-v1. In reinforcement learning area, it is a classic control problem. The agent's job in this environment is to balance a pole that is attached to a cart by pulling the cart to the left or the right. Every time the pole stays upright, the agent is rewarded with +1, and the episode ends if the pole bends too much or the cart moves too far outside of a predetermined area. If 500 seconds have passed, the game is deemed a success [4].

### B. Double DQN (DDQN)

Double DQN's architecture is comparable to that of DQN [5]. It is made up of a deep neural network, which receives the current state as input and returns the estimated Q-values for every conceivable action [6]. However, to be able to fix the overestimation bias issue with traditional Q-learning, DDQN introduces the idea of target networks. The weights from the online network are periodically updated on the target network, a different network. As a result, the learning process is stabilized and the overestimation of action values is reduced.

### C. Deep Recurrent Q-Network (DRQN)

Deep Recurrent Q-Network (DRQN) is an extension of DQN that incorporates recurrent neural networks (RNNs) to capture temporal dependencies in sequential decision-making problems [7]. In DRQN, the input to the network includes not only the current state but also a sequence of previous states. This allows the agent to consider the history of observations and make decisions based on the temporal context. The recurrent units in the network enable the model to retain memory and learn long-term dependencies.

### D. Dueling DQN

The Dueling DQN algorithm [8] separates the estimation of the state value function from the dominance function. The dominance function measures the relative importance of each action taken in that state, while the state value function represents the value present in each state. Dueling DQN is able to increase learning effectiveness and more accurately assess the value of state-action pairs by decoupling these two components.

### E. Knowledge Distillation

The distillation principle employed in this study involves using a pre-trained teacher model to guide the training of a smaller student model [9]. The teacher model, which is a more complex and high-performing DRL model, provides soft targets in addition to the usual rewards during the training process. These soft targets are generated by the teacher model's softmax outputs, which represent the probabilities of actions rather than the hard action choices. The additional outputs are defined using a temperature T in order to "soften" the softmax outputs [10].

$$q_i^{(t)} = \frac{\exp(z_i^{(t)}/T)}{\sum_j \exp(z_j^{(t)}/T)} \quad (1)$$

The distillation's intended objective, the student model, is trained using a mixture of the initial incentives and the soft targets that the teaching model offers. By leveraging the knowledge of the teacher model, the student model can learn more efficiently and generalize better.

Several other papers have explored knowledge distillation in the context of reinforcement learning. "Dual Policy Distillation" [11] presents a method that distills policy knowledge from an ensemble of teacher policies to train a student policy. "Catastrophic Interference in Reinforcement Learning"[12] discusses the several problem of catastrophic interference and proposes a method to mitigate it using distillation. "Distilling Reinforcement Learning Tricks" [13] in Video Game domain illustrate how to make reward shaping, curriculum learning, and splitting a large task into smaller chunks more effectively.

In this study, the distillation technique is applied to the DQN, DDQN, DRQN, and Dueling DQN algorithms. The specific distillation process and its impact on the performance of the models will be further described in subsequent sections.

## III. RESULTS

### A. Experimental Design and Training Details

To ensure that the main difference between training DRL using the environment and training DRL using the teacher's network is that they use different loss functions, the environment setup is the same, including the reward rules, the action selector, and the expected reward function. In DRL, only two things can be changed: the network and the loss function. In the distillation part, the teacher network is based on the DQN/DDQN/Dueling DQN algorithm, while the student network is a smaller-layered version of the DQN/DDQN/Dueling DQN algorithm. In DQN, DDQN, DuelingDQN, and DRQN, students' corresponding compression ratios for the total number of parameters are 25.34%.

In Knowledge Distillation part, the temperature was not used as a variable in this experiment, so it defaulted to 5. Firstly, two networks are utilized in DQN: the teacher model, denoted as Teacher = {n_observations, 128, 128, n_action}, and the student model, denoted as Student = {n_observations, 64, 64, n_action}. The network architectures are given as a sequence of layer sizes. Each architecture starts with an input size of n_observations and consists of hidden layers with sizes 128, 128, and n_action as output size for the teacher model, and 64, 64, and n_action as output size for the student model. Compared to DQN, DRQN (Deep Recurrent Q Networks) includes an LSTM (Long Short-Term Memory) layer in the experiment. The LSTM layer enables DRQN to capture temporal dependencies by incorporating recursive connections to process sequential information in the input observations. The DuelingDQN architecture introduces a separation between the value stream and the advantage stream. It estimates the value function and the benefits of each action using two different paths. The Q-values are then computed by combining the value and advantage streams, allowing for a better representation of the value and action preferences separately.

### B. Evaluation Indexes

The experiment utilizes several quantitative metrics to assess the performance of the distilled post-reinforcement learning algorithm. First, the experiment measures the effectiveness of the algorithm by recording the number of episodes required to reach an average score of 300, which indicates the speed of learning and convergence. In addition, the experiment calculates the average score obtained for the 500th episode as a measure of the overall performance of the algorithm. Finally, by tracking the number of rounds it takes for the score to reach 500 for the first time, the experiment provides insight into the initial progress and learning capabilities of the algorithm. These quantitative metrics provide valuable indicators for evaluating the effectiveness and efficiency of distilled reinforcement learning algorithms on a given task.

## C. Performance Comparison

After comparing different Deep Reinforcement Learning (DRL) algorithms, this experiment found that in relatively simple discrete environments, both DRQN and Dueling DQN consistently outperformed DQN, both before and after distillation. The performances are demonstrated in Table I, II, and III respectively. However, Double DQN showed slower convergence compared to DQN before distillation yet demonstrated superior performance after distillation. These results can be attributed to the following factors:

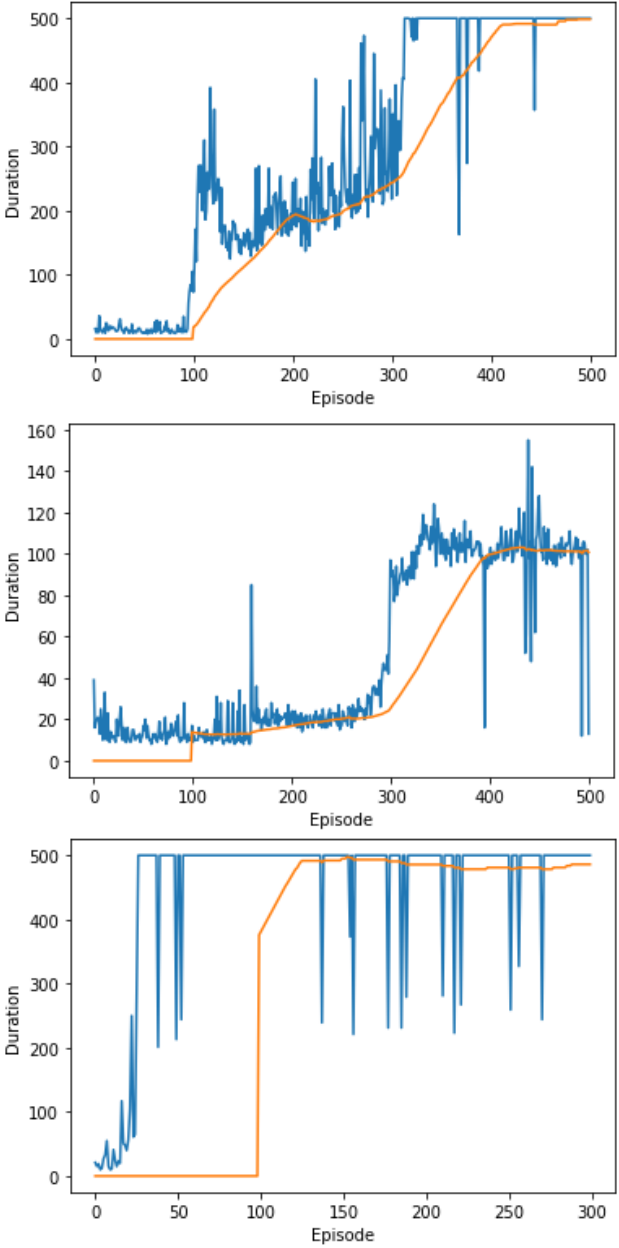

Figure 1. Teacher, student, distillation of DRQN in catpole-v1 (Figure credit: Original)

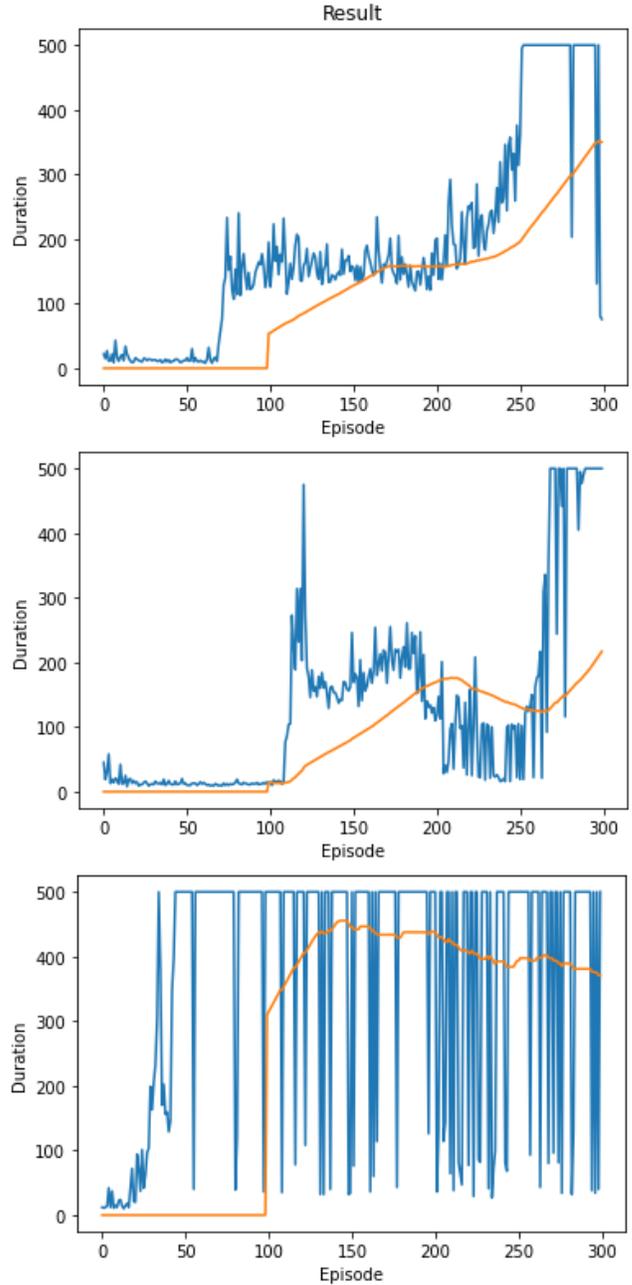

Figure 2. Teacher, student, distillation of Dueling DQN in catpole-v1 (Figure credit: Original)

DRQN and Dueling DQN Advantages: DRQN incorporates recurrent neural networks (RNNs), allowing it to capture temporal dependencies in sequential data, which can be beneficial in certain environments. Dueling DQN utilizes a separate value and advantage function, enabling more accurate value estimation and improved learning. These advantages likely contribute to their superior performance over DQN in the experiment. The improved learning speed of DRQN and Dueling DQN and their rewards are shown in Figure 1 and Figure 2, respectively.

DoubleDQN Convergence Speed: The slower convergence of doubleDQN compared to DQN before distillation can be attributed to the inherent challenges in addressing the overestimation bias. While doubleDQN attempts to mitigate this bias by employing a target network, it may still encounter difficulties in effectively balancing exploration and exploitation. This can result in slower convergence during training.

Distillation Benefits: The improved performance of doubleDQN after distillation can be attributed to the knowledge transfer process from the DDQN teacher network. Distillation can help doubleDQN algorithms learn more effectively and efficiently from the teacher network, which has richer and more complex knowledge. The distillation process can provide a stronger initial policy for doubleDQN by transferring the knowledge from the teacher network to the student network, and guiding the learning process of doubleDQN. This knowledge transfer likely contributes to the enhanced performance observed after distillation.

In summary, the experimental results indicate that DRQN and Dueling DQN outperform DQN in austere discrete environments, both before and after distillation. DoubleDQN exhibits slower convergence before distillation but shows improved performance after distillation compared to DQN. These results underline the merits of particular architectural alterations in DRL algorithms as well as the advantages of knowledge distillation in improving DRL algorithm performance.

TABLE I. PERFORMANCE OF TEACHER MODELS

| Indicators | DQN | DRQN | Double DQN | Dueling DQN |
|---|---|---|---|---|
| episode_num (Avg score=300) | 375 | 326 | — | 283 |
| Average score (500 episodes) | 500 | 500 | 389 | 500 |
| episode_num (Score=500) | 322 | 317 | 140 | 250 |

TABLE II. PERFORMANCE OF STUDENT MODELS

| Indicators | DQN | DRQN | Double DQN | Dueling DQN |
|---|---|---|---|---|
| episode_num (Avg score=300) | — | — | — | 377 |
| Average score (500 episode) | 107 | 97 | 100 | 500 |
| episode_num (Score=500) | — | — | — | 140 |

TABLE III. PERFORMANCE OF DISTILLATED MODELS

| Indicators | DQN | DRQN | Double DQN | Dueling DQN |
|---|---|---|---|---|
| episode_num (Avg score=300) | 200 | 28 | 43 | 27 |
| Average score (500 episode) | 500 | 500 | 500 | 500 |
| episode_num (Score=500) | 150 | 32 | 22 | 31 |

IV. CONCLUSION

The results of this experiment prove that the performance of compressing the original Deep Reinforcement Learning model by knowledge distillation far exceeds the performance of compressing the original DRL model by changing the parameters. Secondly, the experiment above compares the performance of DQN, DRQN, Double DQN and Dueling DQN algorithms after distillation with the original teacher model through the simple environment of discrete-action catpole-v1. It is found that the learning speed and convergence speed of the distilled DRL model are greatly improved. The experiment also demonstrated the feasibility and effectiveness of knowledge refinement in deep reinforcement learning, providing new ideas for actors to critique such algorithms with poor convergence, while training DRL models with fewer parameters but not weaker performance than large models, a trait that can do well with less GPU resources, learn faster, and make judgments more quickly under challenging circumstances. Further research and exploration in this area will likely lead to more efficient and effective reinforcement learning algorithms in resource-constrained environment.